\documentclass[letterpaper]{article} 
\usepackage{aaai24}  
\usepackage{times}  
\usepackage{helvet}  
\usepackage{courier}  
\usepackage[hyphens]{url}  
\usepackage{graphicx} 
\usepackage{natbib}  
\usepackage{caption} 
\usepackage[dvipsnames]{xcolor}
\usepackage{multirow}
\usepackage{footmisc}
\usepackage{svg}
\usepackage{fancyvrb}
\usepackage{circuitikz}
\usepackage{moresize}
\usepackage{subcaption}

\usepackage{algorithm}
\usepackage{algorithmic}

\usepackage{newfloat}
\usepackage{listings}
\DeclareCaptionStyle{ruled}{labelfont=normalfont,labelsep=colon,strut=off} 
\floatstyle{ruled}
\newfloat{listing}{tb}{lst}{}
\floatname{listing}{Listing}

\newcommand{\gpt}{\texttt{GPT-3.5} }
\newcommand{\citeintext}[1]{\citeauthor{#1} \shortcite{#1}}

\title{How Teachers Can Use Large Language Models and Bloom's Taxonomy \\ to Create Educational Quizzes}
\author {
    Sabina Elkins\textsuperscript{\rm 1,\rm 2},
    Ekaterina Kochmar\textsuperscript{\rm 2, \rm 3},
    Jackie C.K. Cheung\textsuperscript{\rm 1, \rm 4}
    Iulian Serban\textsuperscript{\rm 2}
}
\affiliations {
    \textsuperscript{\rm 1}McGill University \& MILA\\
    \textsuperscript{\rm 2}Korbit Technologies Inc.\\
    \textsuperscript{\rm 3}MBZUAI\\
    \textsuperscript{\rm 4}Canada CIFAR AI Chair\\
}

\begin{document}

\maketitle

\begin{abstract} \begin{quote} Question generation (QG) is a natural language processing task with an abundance of potential benefits and use cases in the educational domain. In order for this potential to be realized, QG systems must be designed and validated with pedagogical needs in mind. However, little research has assessed or designed QG approaches with the input from real teachers or students. This paper applies a large language model-based QG approach where questions are generated with learning goals derived from Bloom's taxonomy. The automatically generated questions are used in multiple experiments designed to assess how teachers use them in practice. The results demonstrate that teachers prefer to write quizzes with automatically generated questions, and that such quizzes have no loss in quality compared to handwritten versions. Further, several metrics indicate that automatically generated questions can even improve the quality of the quizzes created, showing the promise for large scale use of QG in the classroom setting.
 \end{quote} \end{abstract}

\section{Introduction}\label{intro} Question generation (QG) is a popular natural language processing (NLP) task. The goal is to generate natural-language questions that are useful and fluent. Many approaches also attempt to generate the corresponding answers, or use the answer to generate the question \cite{kurdi2020systematic,mulla2023automatic}. Due to their recent success in NLP, recent QG research has been dominated by the use of Transformer-based large language models (LLMs) \cite{kurdi2020systematic,liu_2023}. 

An obvious use case for QG is educational applications. A robust QG system could, for example, reduce the time spent by teachers to create educational content such as homework, quizzes, tests, in-class learning activities, and more. Alternatively, it could serve as a practice tool for students. The range of potential uses for educational question generation (EQG) is expansive, especially considering the recent success of LLMs \cite{kasneci2023chatgpt,kurdi2020systematic}.

Unfortunately, there is minimal documented real-world deployment of such systems \cite{kasneci2023chatgpt,kurdi2020systematic}. 
Potential reasons for this lack of adoption may include poor performance of older approaches, rigidity of the systems, and mistrust of users. \citeintext{wang2022needs} conduct a need-finding study with the aim to explore why QG systems are not being used in classrooms. One of their key findings is that QG systems must meet the needs of the educators who are using them in order to be effective and adopted. To achieve this, it is imperative that research in developing educational QG systems takes into account the opinions of their end users.

\begin{figure}[!t]
    \begin{subfigure}[b]{.5\textwidth}
    \begin{lstlisting}[frame=single]
Generate questions in each level of Bloom's
taxonomy.

Passage: {example_context}
Remembering = {example_question}
Understanding = {example_question}
Applying = {example_question}
Analyzing = {example_question}
Evaluating = {example_question}
Creating = {example_question}

Passage: {context}
Remembering =
    \end{lstlisting}
    \caption{\textit{Controlled} prompt template. Due to space limits, a one-shot template is demonstrated here. In actuality, five examples (with different contexts and questions) are used.}
    \end{subfigure}
    \hfill
    \begin{subfigure}[b]{.5\textwidth}
\begin{lstlisting}[backgroundcolor = \color{lightgray}]
Candidates generated using a context about
convergent evolution:
\end{lstlisting}
\begin{lstlisting}[backgroundcolor=\color{SpringGreen}]
Remembering: What is convergent evolution?

Applying: Can you provide an example of 
convergent evolution?

Analyzing: What is the difference between 
analogous and homologous structures or traits?
\end{lstlisting}
        \caption{\textit{Controlled} prompting strategy generation examples.\footref{repo_footnote} All 6 levels of Bloom's taxonomy have generations for each input context, omitted here to save space.}
    \end{subfigure}
    
    \caption{\textit{Controlled} prompting strategy.}
    \label{fig:control_prompt}
\end{figure}

Given the success of LLMs in other tasks, our hypothesis is that they can generate different types of questions from a given context that teachers find useful for creating a quiz with quality comparable to a handwritten version.
Further, we predict that teachers will find the generated candidates more useful when they are generated to correspond with the levels of Bloom’s taxonomy \cite{krathwohl_2002}. Figure \ref{fig:control_prompt} depicts our few-shot prompting strategy to generate educational questions corresponding to Bloom’s taxonomy (more details about this approach can be found in Section \ref{methods}).
In order to evaluate our predictions, we conducted quiz writing experiments designed to compare and contrast three different quiz types.
Multiple aspects of `usefulness' of a quiz and quiz writing approach are considered, including {\em the quality of the resulting quiz}, {\em the efficiency} (w.r.t. time), and {\em the teacher's preferences}.
Our results show that the three types of quizzes are of similar quality. Some metrics even demonstrate an improvement in quality when automatically generated questions are used. We also find that teachers have a strong preference for writing quizzes with the help of automatically generated questions corresponding to Bloom's taxonomy. These results demonstrate the huge potential of EQG for real-world classrooms and the importance of considering the needs of teachers when designing an EQG approach.
\section{Background}\label{background} Recent research in EQG, and QG more generally, revolves around the use of Transformer-based LLMs. These LLMs are deep learning models trained on massive corpora of data to improve their generative performance \cite{zhang_2022}. The reason for applying this approach in QG research is in large part due to its significant performance improvements over earlier rule-based and other types of systems \cite{kurdi2020systematic,steuer2021linguistic,mulla2023automatic}.

The typical training goal for Transformer-based LLMs is next-token prediction, meaning that they learn to predict a probable completion to an initial input text. 
Recent models have begun to also include reinforcement learning in their training procedure. This is the case for \texttt{GPT-3.5}, which is the LLM used in the experiments reported in this paper. Fine-tuning with reinforcement learning from human feedback allows \gpt to outperform its predecessors in the GPT family \cite{ouyang_2022}.

Aligning with the common LLM training objective of the next-token prediction, the emerging paradigm for QG is to feed a textual input, called a {\em prompt}, to an LLM for the model to complete \cite{mulla2023automatic}. 
Designing this prompt to generate a desired output can be a difficult task, which has resulted in a new research direction called {\em prompt engineering}.
One of the most common approaches to prompt engineering involves prepending a string to the context given to an LLM for generation, which is called a {\em prefix style prompt} \cite{liu_2023}. For instance, say a machine learning teacher wished to generate questions about gradient descent. A simple strategy they could apply is to prompt an LLM with the input:
\begin{quote}
    \textit{Generate a question about gradient descent.}
\end{quote}
To increase the specificity of the generated questions, the teacher could provide more context. For example, they might instead write a prompt with a textbook passage about a specific aspect of gradient descent, such as:
\begin{quote}
    \textit{Generate a question \underline{from the following passage}: $<$...$>$}
\end{quote}
To further control generation, the teacher's input could contain a \textit{control element} – a keyword that will guide the generation \cite{mulla2023automatic}. For instance, they could prompt an LLM with:
\begin{quote}
    \textit{Generate a \underline{multiple-choice question} from the following passage: $<$...$>$}
\end{quote}
In this simple example, we have presented three different prompts, all of which can produce different questions. Adding in different word choices and other strategies for controlling the generation quickly makes prompt engineering a complex problem to optimize.

Another aspect in prompt engineering is the inclusion of examples of the desired output format and style within the prompt itself. This is often called {\em few-shot learning}. In brief, few-shot learning prompts consist of an instruction, a few examples, and the task at hand. The examples are used to adapt LLMs to unseen scenarios without additional training or fine-tuning \cite{liu_2023}. For instance, following the earlier example, a teacher might prompt an LLM to generate a true-or-false type of question by including examples of such  questions. 

The ability to add more specificity to educational QG has allowed researchers to generate questions at different levels of difficulty, with different pedagogical goals in mind, and more.
For example, \citeintext{wang2022towards} try a collection of different prompting strategies in an effort to optimize educational QG. They conclude that using shorter input contexts and few-shot learning results in higher quality candidate questions.
Recently, \citeintext{elkins2023useful} have demonstrated how to generate questions at different levels of question taxonomies, which are organizational structures taken from pedagogical literature, such as Bloom's taxonomy of learning goals.
The authors demonstrate successful generation of questions at various complexity levels and with different learning goals.

Despite these exciting recent research developments with educational QG, there are only a few documented cases of these techniques being used in real-world classrooms \cite{kasneci2023chatgpt,kurdi2020systematic}.
As previously mentioned, \citeintext{wang2022needs} find that the lack of alignment between research goals and what teachers actually need and want from EQG is the reason that such systems are not deployed in real-world classrooms.
Prior work aiming to explore the needs, opinions, and attitude of teachers and students towards using automatically generated content is few and far between. Nevertheless, there exist a few relevant precedent papers:
\begin{itemize}
    \item The work by \citeintext{van_2022} outlines a NLP system for translating textbooks into interactive courseware. The authors conduct a large user study which shows equal student performance on machine-generated and human-written questions. However, their QG system is mostly rule-based, and their generated questions are relatively simple (i.e., concept-matching and fill-in-the-blank).
    \item The aforementioned work by \citeintext{elkins2023useful} evaluated their generated candidate questions with real teachers in an effort to more accurately assess their pedagogical usefulness. They find that their generated questions are highly rated by teachers. However, this work only evaluates questions at the individual question-level rather than at a quiz-level.
    \item The work by \citeintext{laban_2022} moves beyond question-level evaluations to a quiz writing task, similar to our work in this paper. The authors design a task where teachers make a quiz exclusively with candidate questions that were automatically generated. The teachers also marked the candidates as acceptable or not as they went; however, the final global acceptance rate was only $52\%$. Thus, while the authors take important steps towards the evaluation of QG in a realistic scenario, their generations themselves appear to leave room for improvement.
\end{itemize}

\section{Methods}\label{methods} The EQG in this work was conducted by prompting \gpt to generate questions from a given input passage. Two different strategies are used in order to compare the pedagogically designed generation approach to a more simplistic approach to QG.
The strategies are referred to as \textit{controlled} and \textit{simple}, respectively.
Individually, both types of candidate generations are supplied to teachers in controlled experiments to assess their usefulness in practice (the details of these experiments are explained in Section \ref{evaluation}). The following subsections will cover details of the input contexts, and the two prefix style prompting strategies used.

\subsection{Contexts}
Candidate educational questions were generated from a set of $24$ passages sourced from Wikipedia, with each passage containing $5$ context paragraphs, $6$ to $9$ sentences in length. The total number of input contexts used to generate questions is, thus, $120$. The length of these contexts was determined by empirical results in preliminary work.
This set of passages was manually gathered. The specific Wikipedia articles were generally chosen via hyperlinks from the domain's main Wikipedia or glossary page in order to ensure they were relevant to the foundations of the given domain. 
Two domains were used: biology (BIO) and machine learning (ML). More than one domain is included in order to take steps towards demonstrating domain-agnostic results.
Each domain has $12$ passages, totaling $60$ input contexts. The contexts underwent minor pre-processing before use: this included removal of citations, hyperlinks, footnotes and phonetic spellings, re-formatting of full sentence bullet-point lists into paragraphs, and other minor data cleaning steps.\footnote{\label{repo_footnote} The input contexts, more details about Bloom's taxonomy, the human authored few-shot examples, all of the generated candidates and quizzes, the annotator demographics, and more can be found at \url{https://anonymous.4open.science/r/EQG_in_practice-2752/README.md}.}

\subsection{\textit{Simple} Prompting Strategy}
The \textit{simple} prompting strategy uses a generic strategy to generate questions with \texttt{GPT-3.5}\footnote{The model \texttt{text-davinci-003} was accessed through the OpenAI API. The dates on which the model was queried are available in the supplementary material.} in an attempt to assess how well the model can generate pedagogically useful questions without any additional prompt engineering. The prompt template can be seen in Figure \ref{fig:simple_prompt}. The number of questions generated at once is six in order to produce the same amount of questions as the \textit{controlled} strategy described below. Empirical results from preliminary experimentation showed that generating all of the questions together produced more diverse outputs, whereas generating them separately produced duplicate questions.


\begin{figure}[!h]
    \begin{subfigure}[b]{.5\textwidth}
    \begin{lstlisting}[frame=single]
Generate 6 questions.

Passage: {context}
Questions:
    \end{lstlisting}
    \caption{\textit{Simple} prompt template.}
    \end{subfigure}
    \hfill
    \begin{subfigure}[b]{.5\textwidth}
\begin{lstlisting}[backgroundcolor = \color{lightgray}]
Candidates generated using a context about
convergent evolution:
\end{lstlisting}
\begin{lstlisting}[backgroundcolor = \color{SpringGreen}]
What is the definition of convergent evolution?

What are some common functions found in bird, 
bat, and pterosaur wings?
\end{lstlisting}
        \caption{\textit{Simple} prompting strategy generation examples.\footref{repo_footnote} There are actually 6 generations, omitted here to save space.}
    \end{subfigure}
    
    \caption{\textit{Simple} prompting strategy.}
    \label{fig:simple_prompt}
\end{figure}

\subsection{\textit{Controlled} Prompting Strategy}
The \textit{controlled} prompting strategy uses a pedagogical question taxonomy to generate questions with different learning goals in mind. Bloom's taxonomy is a popular framework for categorizing learning objectives in educational materials \cite{krathwohl_2002}.
The taxonomy contains six levels of learning, arranged in a hierarchical order from `lower'- to `higher'-level thinking skills: \textit{remembering}, \textit{understanding}, \textit{applying}, \textit{analyzing}, \textit{evaluating}, and \textit{creating}.\footref{repo_footnote} Bloom's taxonomy helps teachers design instructional content that targets specific learning goals.

The generation strategy used in this paper is designed to reduce the overlap between generated candidates. Rather than generating one question and taxonomic level at a time, all six questions for a given context are generated at once, as shown in Figure \ref{fig:control_prompt}. In preliminary experimentation, this approach empirically showed a reduction in the generation of identical questions and a greater diversity and adherence to the taxonomic levels.

The \textit{controlled} prompting strategy also uses few-shot learning. Following success with preliminary experimentation, five-shot learning (i.e., five examples within the prompt) is used. As seen in Figure \ref{fig:control_prompt}, a single example contains a prompt and six questions, one for each level of Bloom's taxonomy. The examples were handcrafted by a domain expert, and reviewed by another expert for both question quality and adherence to the intended taxonomic level. Each domain had five sets of contexts and examples, totaling $10$ contexts and $60$ questions.\footref{repo_footnote}

\section{Evaluation}\label{evaluation} In order to assess the applicability of our two varieties of automatically generated questions to a real-world academic setting, we must actually ask real teachers to use them. Therefore, we designed an experiment to replicate a teacher's creation of a reading quiz. We wanted to compare the resulting quizzes' quality when teachers are writing quizzes from scratch and when they have access to automatically generated questions while writing the quiz.
Section \ref{quiz_quality_methods} introduces the measures of quiz quality considered in this paper. Then, Section \ref{quiz_writing_methods} explains the quiz writing experiments conducted by real teachers. 

\subsection{Quiz Quality}\label{quiz_quality_methods}
To be able to compare the quality of questions and quizzes written with the aid of \gpt to handwritten ones, we must first establish how to measure the quality of a quiz in the first place. However, the quality of a quiz is often a subjective metric. Teachers have differing opinions on what makes a good quiz based on their individual teaching styles. We defined a set of metrics to measure individual aspects of a quiz that capture different aspects of its quality, with only \textit{usefulness} designed to purely reflect the annotator's opinion.

Based on previous research, we identified the following four quiz-level metrics to include.\footnote{The ordinal metrics also have definitions for each category available in the supplementary material.} 
A good reading quiz will be relevant to the teaching material. It will be natural and coherent, meaning that it will not confuse a student taking it. Above all, a good quiz will be approved by a teacher for classroom use. The following metrics attempt to assess each of these aspects:
\begin{itemize}
    \item \textit{Coverage} is a numerical metric in $[0,1]$ which measures how much of the input passage is reflected in the final quiz. To measure this, we mapped each question to the sentences in the passage that contained pertinent information to the answer. For consistency, all questions are mapped to any possible answer in the text (i.e., both instances of repeated information are selected, open-ended questions select more of the text, etc.). A \textit{coverage} ratio is then calculated from the length of the mapped text and the whole passage's length. This metric is inspired by the pyramid method for annotating summaries, which uses a similar strategy \cite{nenkova2004evaluating}.
    
    \item \textit{Structure} is an ordinal metric from $1$ to $3$ which measures whether the set of questions make sense together. In other words, if they are intuitively linked together with a natural/understandable flow (e.g., from easy to difficult, or from start to finish of the context).
    Previously, a similar metric has been used for conversational QG where questions must be logically linked in order for a conversation to be natural \cite{mulla2023automatic}.
    
    \item \textit{Redundancy} is an ordinal metric from $1$ to $3$ which measures if there is redundancy/repetition within the quiz, for example if there are two questions that ask for the same answer without any different perspective or thought process required from the student.
    Previously, a similar metric has been used for conversational QG where questions must not be repetitive in order for a conversation to be natural \cite{mulla2023automatic}.

    \item \textit{Usefulness} is an ordinal metric from $1$ to $4$ which measures if a teacher would use the quiz in an assessment they create for their own class. 
    Note that the quiz does not necessarily need to be entirely answerable from the context in order to be considered useful. Previously, similar ordinal metrics have been used in \citeintext{elkins2023useful}, and on different scales in \citeintext{steuer2021linguistic} and \citeintext{mulla2023automatic}.
\end{itemize}

Beyond the quality of the whole quiz, it is important to ensure that the individual questions are also of high quality. Thus we also outlined three question-level metrics, meaning they are evaluated for each question in a quiz. This is not an exhaustive evaluation, but the following covers basic aspects of a question's quality:
\begin{itemize}
    \item \textit{Relevancy} is a binary metric which measures whether the question is semantically relevant to the input context. Previously, similar binary approaches have been used in \citeintext{steuer2021linguistic} and \citeintext{elkins2023useful}, and on different scales by \citeintext{mulla2023automatic}.
    
    \item \textit{Fluency} is a binary metric which measures whether the set of questions are grammatically correct and use clear language. Similarly to the previous metric, previous approaches by \citeintext{mazidi2014pedagogical} and \citeintext{elkins2023useful} have applied this binary metric, and it has been used more generally on different scales in \citeintext{mulla2023automatic}.
    
    \item \textit{Answerability} is a binary metric which measures whether the question can be answered from the input context. It is not necessary to be able to find a passage from the input that is an answer to the question; it is enough if a student could reasonably answer the question from the context (for example, applying logic explained in the passage to a new situation makes the question ‘answerable’). As above, previous work by \citeintext{steuer2021linguistic} and \citeintext{elkins2023useful} uses a similar binary metric, and \citeintext{mulla2023automatic} suggests similar metrics on different scales.
\end{itemize}

\subsection{Quiz Writing Experiments}\label{quiz_writing_methods}
The quiz writing experiments were designed to mimic a teacher's creation of a reading quiz. In order to be able to measure and compare teacher's quiz writing processes, the setting was tightly controlled. Discussion of the potential limitations this may introduce can be found in Section \ref{limitations}. Before conducting the experiments, a pilot with four teachers in the ML domain was conducted to ensure the metrics and annotator training were unambiguous. This pilot resulted in minor changes to the wording in the training and metric definitions, but no major experimental flaws were discovered.

There were $24$ quiz writing teachers, $12$ in each domain. The BIO teachers were found through the freelance platform Upwork and have at least a high-school level of teaching/tutoring experience. The ML teachers were recruited through word-of-mouth at the institutions of the first author and have at least a university level of teaching/tutoring experience. All teachers were fairly monetarily compensated and signed a consent form before participating. They are all proficient in the English language, and are from relatively diverse demographics.\footref{repo_footnote} The teachers completed a training module where they were provided with example quizzes in their domain and were guided through the creation of each quiz type to ensure they understood the task at hand.

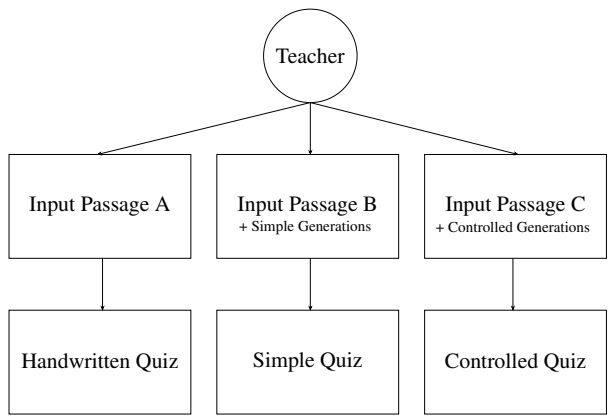
\begin{figure}[!ht]
    \centering
    \resizebox{0.45\textwidth}{!}{
    \begin{circuitikz}
        \tikzstyle{every node}=[font=\HUGE]
        \node [font=\Huge] at (6.75,16.5) {+ Simple Generations};
        \node [font=\Huge] at (16.75,16.5) {+ Controlled Generations};
        \draw [, line width=1pt ] (-7.5,20) rectangle  node {Input Passage A} (1.25,15);
        \draw [, line width=1pt ] (2.5,20) rectangle  node {Input Passage B} (11.25,15);
        \draw [, line width=1pt ] (12.5,20) rectangle  node {Input Passage C} (21.25,15);
        \draw [, line width=1pt ] (7,24.75) circle (2.25cm) node {Teacher} ;
        \draw [ line width=0.5pt, -Stealth] (7,22.5) -- (7,20);
        \draw [ line width=0.5pt, -Stealth] (7,22.5) -- (17,20);
        \draw [ line width=0.5pt, -Stealth] (7,22.5) -- (-3.25,20);
        \draw [, line width=1pt ] (-7.5,12.5) rectangle  node {Handwritten Quiz} (1.25,7.5);
        \draw [, line width=1pt ] (2.5,12.5) rectangle  node {Simple Quiz} (11.25,7.5);
        \draw [, line width=1pt ] (12.5,12.5) rectangle  node {Controlled Quiz} (21.25,7.5);
        \draw [ line width=0.5pt, -Stealth] (-3,15) -- (-3,12.5);
        \draw [ line width=0.5pt, -Stealth] (7,15) -- (7,12.5);
        \draw [ line width=0.5pt, -Stealth] (16.75,15) -- (16.75,12.5);
    \end{circuitikz}
    }
    \caption{Quiz writing experiment diagram depicting the three quiz writing settings each teacher completed.}
    \label{fig:experiment_diagram}
\end{figure}

The process a single teacher underwent, irrespective of their domain, is depicted in Figure \ref{fig:experiment_diagram}. Each teacher wrote three quizzes, between five and ten questions in length. With $12$ teachers and $12$ passages per domain, each passage was used to create each type of quiz. Note that each teacher received three different passages, to reduce the potential bias due to a teacher working with material they have already seen.
The three quiz types were: \textit{handwritten}, \textit{simple}, and \textit{controlled}. To create a \textit{handwritten quiz}, the teacher simply read the passage and wrote a quiz from scratch. To create a \textit{simple quiz}, the teacher read the passage and related candidates generated with the \textit{simple} prompting strategy. They then created a quiz with the freedom to copy generated questions directly, copy and alter them, or write questions from scratch. Similarly, to create a \textit{controlled quiz}, the teacher read the passage and related \textit{controlled} candidates, and then wrote a quiz.
The quizzes were written in a random order to reduce any potential for biased results from the ordering of the quiz writing subtasks. 

The teachers were asked to record their screen during the quiz writing process. These videos were analyzed to assess teachers' experiences writing each kind of quiz. The \textit{time} taken to write the quiz was measured, including how long the reading of the passage and additional candidates took. The \textit{length} of the final quiz was measured, with a minimum of five and a maximum of $10$ questions. And finally, the \textit{source} of the questions was noted. In other words, it was recorded if the question was directly copied from the \gpt generations, if it was copied and altered by the teacher, or if it was written completely from scratch.

Upon completion of the experiment, the teachers were asked to complete a brief post-quiz to better understand their feedback on the three quiz writing tasks. There was a section for free-form comments about the experiment. Then, the teachers were asked to pick which of the three quiz types was their preferred type and provide their reasoning.

Finally, eight other annotators (four per domain) analyzed the quality of the resulting quizzes using the metrics introduced in Section \ref{quiz_quality_methods}. They were recruited in the same way as outlined above, and have similar teaching experience and English proficiency.
Six quizzes were seen by all four of the annotators in each domain in order to measure inter-annotator agreement. The rest of the quizzes were seen by two annotators. Having more than one annotator evaluate each quiz enables a more robust measurement of the quizzes' quality. The annotators were not informed which quizzes included automatically generated questions. 
\section{Results}\label{results} This section covers the key results for the evaluations of quiz quality and the quiz writing process. First, Section \ref{quiz_quality_results} will compare and contrast the measured quality of the three different quiz types. Then, Section \ref{quiz_writing_results} will discuss the teacher's experiences in the different quiz writing settings.

\subsection{Quiz Quality}\label{quiz_quality_results}
Overall, the results from the quiz quality evaluations show that there is no notable loss of quality between the \textit{handwritten} and the two generation quiz types. There is even an argument that an increase in quiz quality can be seen in some of the following results.

The three question-level metrics demonstrate a low count of \textit{irrelevant}, \textit{disfluent}, and \textit{unanswerable} questions across all cohorts and quiz types. When comparing across quiz types, these results are relatively consistent. This implies that the use of generated candidates does not significantly increase or decrease the quality of quiz questions along these three aspects. The annotator agreement on these metrics is `fair', with an average pairwise Cohen's $\kappa$ value of $0.3$ for the BIO cohort and $0.6$ for the ML cohort~\cite{landis1977application}. \footnote{Due to low percentages of \textit{irrelevant}, \textit{disfluent}, and \textit{unanswerable} questions the agreement values are calculated on a unbalanced dataset and may not be an accurate representation of the real agreement on problem cases.}
\begin{table}[ht!]
    \centering
    \begin{tabular}{|c|cc|cc|cc|}
    \hline
    \multirow{2}{*}{{\bf Metric}} & \multicolumn{2}{c|}{{\bf Handwritten}} & \multicolumn{2}{c|}{{\bf Simple}}      & \multicolumn{2}{c|}{{\bf Controlled}}  \\ \cline{2-7} 
                            & \multicolumn{1}{c|}{{\bf BIO}}  & {\bf ML}   & \multicolumn{1}{c|}{{\bf BIO}}  & {\bf ML}   & \multicolumn{1}{c|}{{\bf BIO}}  & {\bf ML}   \\ \hline
    {\bf Irrelevant}              & \multicolumn{1}{c|}{0.03} & 0.00 & \multicolumn{1}{c|}{0.05} & 0.05 & \multicolumn{1}{c|}{0.04} & 0.00 \\ \hline
    {\bf Disfluent}               & \multicolumn{1}{c|}{0.14} & 0.11 & \multicolumn{1}{c|}{0.09} & 0.08 & \multicolumn{1}{c|}{0.14} & 0.11 \\ \hline
    \begin{tabular}[c]{@{}c@{}}{\bf Unanswer-}\\ {\bf able}\end{tabular} & \multicolumn{1}{c|}{0.08} & 0.10 & \multicolumn{1}{c|}{0.10} & 0.08 & \multicolumn{1}{c|}{0.12} & 0.10 \\ \hline
    \end{tabular}
    \caption{Question-level quality metrics by cohort and quiz type. The values are the proportion of questions that are \textit{irrelevant}, \textit{disfluent}, or \textit{unanswerable}.}
\end{table}

\begin{figure}[!h]
     \centering
     \includegraphics[width=0.49\textwidth]
     {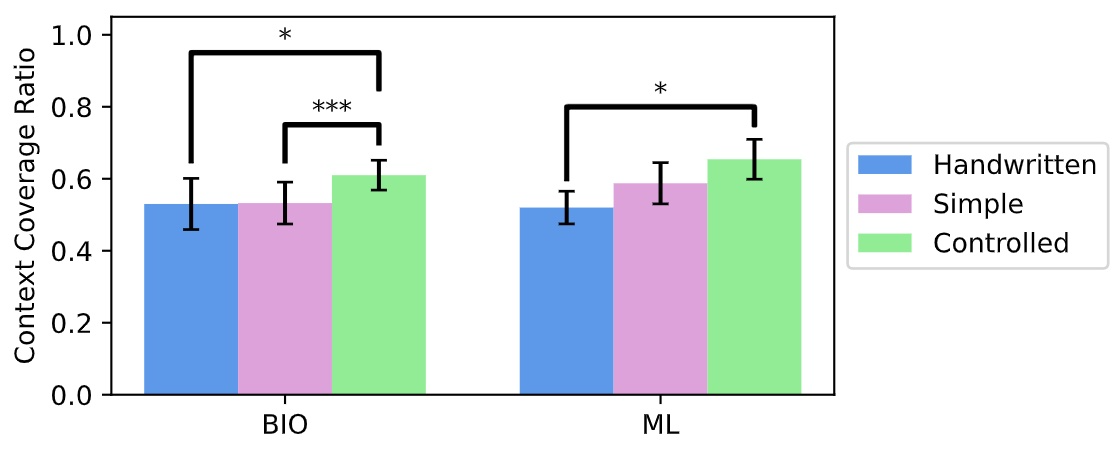}
     \caption{Context coverage by cohort and quiz type. * represents a significant difference at the $\alpha = 0.05$ level, and *** represents a significant difference at the $\alpha = 0.001$ level. The error bars represent $95\%$ confidence intervals.}
     \label{fig:coverage_by_quiz_type}
\end{figure}

The \textit{coverage} results show significant improvement when teachers utilize generated questions during their quiz writing process.\footnote{N.b., a single \textit{coverage} value for each quiz was annotated by the first author of this paper, so no measurement of agreement is available for this metric. We leave this for future work.} Figure \ref{fig:coverage_by_quiz_type} depicts the increasing ratio of input text that is covered between \textit{handwritten} and \textit{simple}, as well as \textit{simple} and \textit{controlled} quiz types. In the ML cohort, a significant difference is observed between the \textit{coverage} ratios in the \textit{handwritten} and \textit{controlled} quizzes. In the BIO cohort, we observe the same significant difference, as well as the difference between the \textit{coverage} ratios in the \textit{simple} and \textit{controlled} quizzes. While it is not the only important aspect of a reading quiz, the increased \textit{coverage} provided by automatically generated questions can benefit teachers in their writing process.

\begin{figure}[h!]
    \centering
    \includegraphics[width=0.49\textwidth]{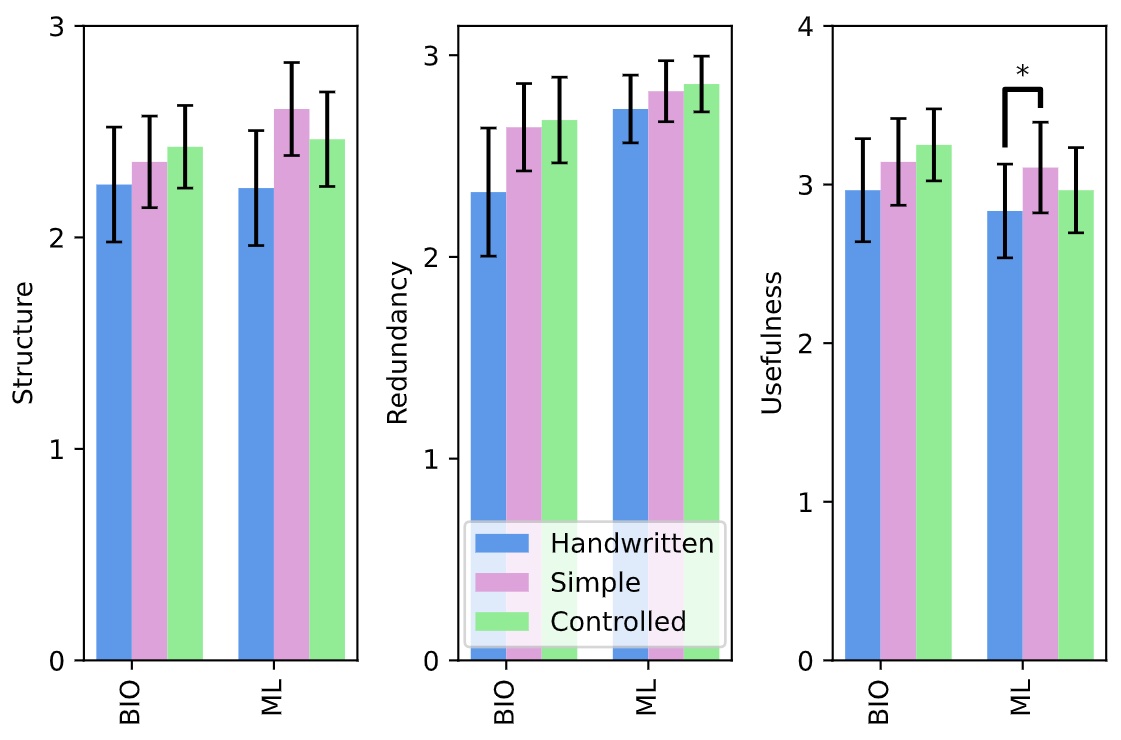}
    \caption{Quiz-level quality metrics. The error bars represent $95\%$ confidence intervals. * represents a significant difference at the $\alpha = 0.05$ level.}
    \label{fig:quiz-level-metrics}
\end{figure}

The three ordinal quiz-level metrics similarly have positive results across the board, as seen in the mean column of Table \ref{tab:quiz_lvl_metrics_agreement}. The annotator agreement values are quite high. These metric's agreement is measured with Kendall's $\tau$ as it is suitable for ordinal scales \cite{schaeffer1956concerning}. All average pairwise values are above $0.5$, and most above $0.81$.
\begin{table}[ht!]
    \centering
    \begin{tabular}{|c|cc|cc|}
    \hline
    \multirow{2}{*}{{\bf Metric}}                                        & \multicolumn{2}{c|}{{\bf Mean}}        & \multicolumn{2}{c|}{{\bf Kendall's $\tau$}} \\ \cline{2-5} 
                                                                   & \multicolumn{1}{c|}{{\bf BIO}}  & {\bf ML}   & \multicolumn{1}{c|}{{\bf BIO}}   & {\bf ML}    \\ \hline
    \begin{tabular}[c]{@{}c@{}}{\bf Structure}\\ {[}1,3{]}\end{tabular}  & \multicolumn{1}{c|}{2.35} & 2.43 & \multicolumn{1}{c|}{0.81}  & 0.83  \\ \hline
    \begin{tabular}[c]{@{}c@{}}{\bf Redundancy}\\ {[}1,3{]}\end{tabular} & \multicolumn{1}{c|}{2.55} & 2.80 & \multicolumn{1}{c|}{0.59}  & 0.80   \\ \hline
    \begin{tabular}[c]{@{}c@{}}{\bf Usefulness}\\ {[}1,4{]}\end{tabular} & \multicolumn{1}{c|}{3.12} & 2.97 & \multicolumn{1}{c|}{0.95} & 0.52 \\ \hline
    \end{tabular}
    \caption{Quiz-level quality metrics and annotator agreement. The \textit{structure} and \textit{redundancy} metrics are on ordinal scales from 1 to 3, and the \textit{usefulness} metric is on an ordinal scale from 1 to 4.} \label{tab:quiz_lvl_metrics_agreement}
\end{table}
Figure \ref{fig:quiz-level-metrics} shows the difference in quiz-level metrics between the three quiz types. Notably, for all three metrics and in both cohorts, one of the generation type quizzes is the highest rated. In the ML cohort there is even a significant difference in the \textit{usefulness} ratings between the \textit{handwritten} and \textit{simple} quizzes.
This points to the fact that the generations can help improve the quality of quizzes. Further optimization of the generation process, perhaps with input from the teachers directly on what types of questions will be useful in a given setting, could show even stronger quality improvements.

\subsection{Quiz Writing Experience}\label{quiz_writing_results}
\begin{figure}[!h]
     \centering
     \includegraphics[width=0.49\textwidth]{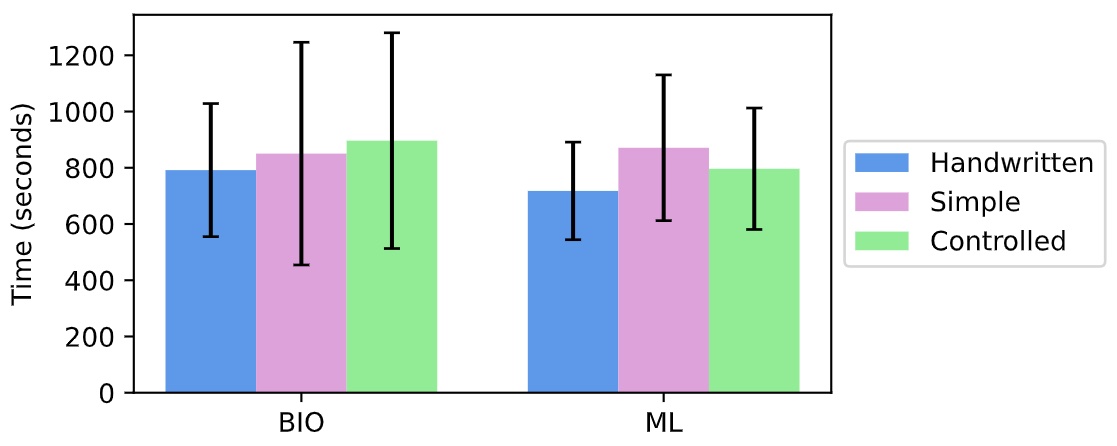}
     \caption{Time taken to create a quiz by cohort and quiz type. The error bars represent $95\%$ confidence intervals.}
     \label{fig:time_by_quiz_type}
\end{figure}

The quiz writing results demonstrate no consistent loss of efficiency when teachers use generated questions, at least with respect to the \textit{time} taken. Figure \ref{fig:time_by_quiz_type} depicts the mean \textit{time} to write each type of quiz for the two cohorts. The mean values are all relatively close to one another, and the 95\% confidence intervals demonstrate that there appears to be no distinct difference in the quiz types with respect to \textit{time}. Further analysis of these results demonstrate that the \textit{time} metric is more dependent on the teacher and the particular passage than the generated candidate questions.

\begin{figure}[!h]
     \centering
     \includegraphics[width=0.49\textwidth]{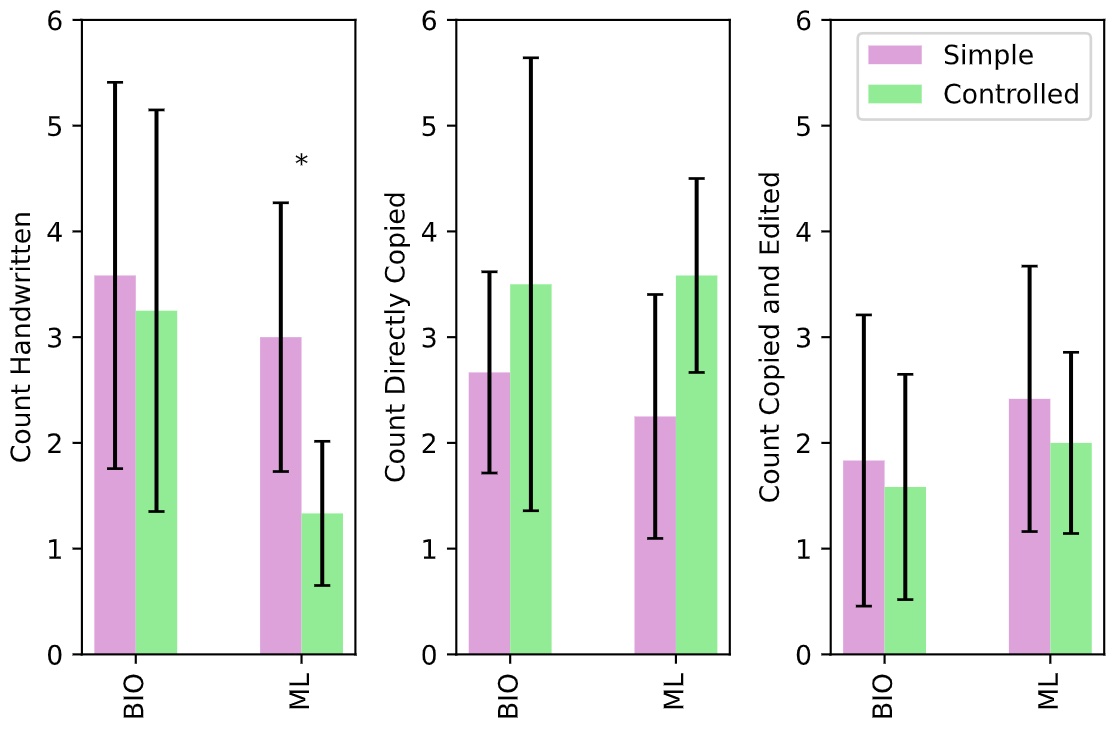}
     \caption{Teachers use of generated questions by cohort. The error bars represent $95\%$ confidence intervals. * represents a significant difference at the $\alpha = 0.05$ level.}
     \label{fig:generated_question_sources_by_cohort}
\end{figure}

The average \textit{length} of the quizzes was $8.14$ in the BIO cohort and $7.11$ in the ML cohort. The mean number of questions in each quiz type was within one question to the mean across all types, demonstrating that the use of generations did not alter this variable.
A more interesting comparison can be seen in Figure \ref{fig:generated_question_sources_by_cohort}, where the \textit{simple} and \textit{controlled} quiz types' question \textit{sources} are compared. In both cohorts, the teachers hand write fewer questions when they have the \textit{controlled} generations at their disposal. In fact, in the ML cohort, this difference is statistically significant. Again in both cohorts, teachers directly copy more of the \textit{controlled} generations into their quizzes than the \textit{simple} generations; and correspondingly they copy and edit fewer questions. This finding demonstrates that teachers choose to use the questions generated with Bloom's taxonomy more than other generations, providing motivation to continue QG research with pedagogical goals in mind.

\begin{figure}[!h]
     \centering
     \includegraphics[width=0.49\textwidth]{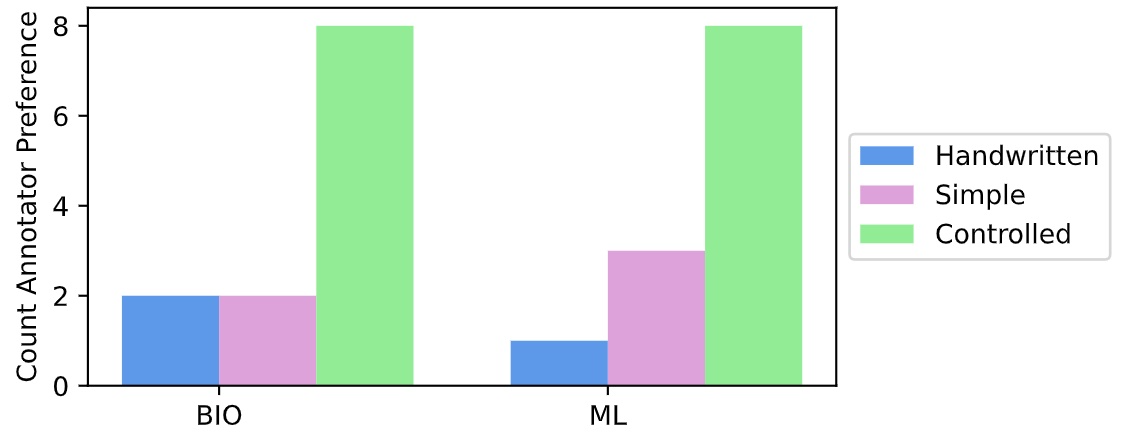}
     \caption{Teacher quiz type preference by cohort.}
     \label{fig:annotator_quiz_type_preference}
\end{figure}

The post-quiz results demonstrate a strong preference from the teachers in both cohorts towards the use of \textit{controlled} generations, as depicted in Figure \ref{fig:annotator_quiz_type_preference}.
The following comments from the teachers provide additional evidence for the fact that they find the \textit{controlled} generations most useful for their quiz writing. Teachers stated that \textit{``The generated questions for each specific type were incredibly useful."} and \textit{``I particularly liked the `creating' questions, as I wouldn't have come up with most of these myself."} Another teacher commented that they \textit{``tried to incorporate different command terms and level questions to prepare a quiz that tests the depth of a student's understanding"}, indicating the intended usage of the questions in different levels of Bloom's taxonomy. This finding corroborates this paper's guiding intuition that QG methods should be designed with pedagogical goals in mind.

\section{Limitations}\label{limitations} It is important to note some limitations of this work.
Firstly, we acknowledge that the quiz writing setting is to some extent contrived. In reality, teachers' quiz writing experiences will be subjective: they might use additional resources or existing knowledge, write a draft and return to it later to edit, not constrain the number of questions, have different learning goals in mind, and more. The quiz writing setting was necessary to control certain variables for comparison within this work, but it creates the possibility that the results do not reflect the reality of quiz writing for teachers. Future work should aim to remove these constraints to better assess how teachers might realistically use generated questions.
Secondly, this work only considers one LLM, two domains, the English-language setting, and a limited number of teachers. Although out of scope for this work, future work should aim to expand the variety of these aspects to assess how generated questions might benefit additional educational settings.
Thirdly, a missing consideration in this work is the other half of the educational setting: students. Future work should include student goals, opinions, and performance in order to more comprehensively understand the implications of using automatically generated questions in the classroom.
Despite these limitations, this work ultimately takes a step in the right direction for the future of QG research within realistic educational use cases.
\section{Conclusion}\label{conclusion} This paper aims to show that LLMs are capable of generating different types of questions from a given context that teachers find useful to create a quiz that is of comparable quality to a handwritten version.
To do this, quiz writing experiments were conducted comparing three types of quizzes: \textit{handwritten}, \textit{simple}, and \textit{controlled} quizzes.
The \textit{controlled} quizzes utilized questions generated to correspond with the levels of Bloom’s taxonomy.
The results demonstrate that teachers strongly prefer writing quizzes with the help of \textit{controlled} generations.
They also directly copy more of the \textit{controlled} generations than the \textit{simple} generations, indicating that these questions are of higher quality or better suited to a teacher's goals. This confirms our hypothesis that teachers find automatically generated pedagogical questions useful for quiz writing.
Additionally, an evaluation of quiz quality showed that the quizzes with both \textit{controlled} and \textit{simple} generations are of comparable quality. Some metrics even point towards their superior quality, when compared to \textit{handwritten} quizzes.
We hope that these findings will help to direct the future of educational QG research towards practical applications that meet the goals and needs of teachers and students.
\section{Acknowledgements}\label{acks} The authors would like to thank Mitacs for their grant for this project, and CIFAR for their continued support. Additional thanks are extended to the teachers and annotators for their time and effort and the anonymous reviewers for their valuable feedback.

\bibliography{references}

\end{document}